\documentclass{article}

\usepackage{arxiv}
\usepackage{multirow,graphicx,enumerate,color,amsfonts,amsthm,amssymb,mathrsfs,dsfont,textcomp,subfigure,array,float,enumitem,xcolor,cancel,soul}
\usepackage[utf8]{inputenc} % allow utf-8 input
\usepackage[T1]{fontenc}    % use 8-bit T1 fonts
\usepackage{hyperref}       % hyperlinks
\usepackage{url}            % simple URL typesetting
\usepackage{booktabs}       % professional-quality tables
\usepackage{amsfonts}       % blackboard math symbols
\usepackage{nicefrac}       % compact symbols for 1/2, etc.
\usepackage{microtype}      % microtypography
\usepackage{lipsum}		% Can be removed after putting your text content
\usepackage{graphicx}
\usepackage{amsmath}

\title{Artificial Neural Network Approach for the Identification of Clove Buds Origin Based on Metabolites Composition}

\author{ \href{http://orcid.org/0000-0001-8331-5793}{\includegraphics[scale=0.06]{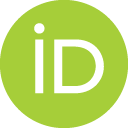}\hspace{1mm}Rustam}\thanks{alternative email: rustamtelu@telkomuniversity.ac.id} \\
Industrial and Financial Mathematics Research Group\\
Faculty of Mathematics and Natural Sciences\\
Institut Teknologi Bandung, Indonesia \\
\texttt{rustam.math@s.itb.ac.id} \\
\And
\href{http://orcid.org/0000-0002-0564-1902}{\includegraphics[scale=0.06]{orcid.png}\hspace{1mm}Agus Y.~Gunawan} \\
Industrial and Financial Mathematics Research Group\\
Faculty of Mathematics and Natural Sciences\\
Institut Teknologi Bandung, Indonesia \\
\And
\href{http://orcid.org/0000-0002-8079-4677}{\includegraphics[scale=0.06]{orcid.png}\hspace{1mm}Made T.~A.~P.~Kresnowati} \\
Food and Biomass Processing Technology Research Group\\
Faculty of Industrial Technology\\
Institut Teknologi Bandung, Indonesia \\
}

\renewcommand{\headeright}{}
\hypersetup{
pdftitle={A template for the arxiv style},
pdfsubject={q-bio.NC, q-bio.QM},
pdfauthor={David S.~Hippocampus, Elias D.~Striatum},
pdfkeywords={First keyword, Second keyword, More},
}

\begin{document}
\maketitle

\begin{abstract}
This paper examines the use of artificial neural network approach in identifying the origin of clove buds based on metabolites composition. Generally, large data sets are critical for accurate identification. Machine learning with large data sets lead to precise identification based on origins. However, clove buds uses small data sets due to lack of metabolites composition and their high cost of extraction. The results show that  backpropagation and resilient propagation with one and two hidden layers identifies clove buds origin accurately. The backpropagation with one hidden layer offers 99.91\% and 99.47\% for training and testing data sets, respectively. The resilient propagation with two hidden layers offers 99.96\% and 97.89\% accuracy for training and testing data sets, respectively.
\end{abstract}

% keywords can be removed
\keywords{Artificial neural networks \and Backpropagation \and Resilient propagation \and Clove buds}

\section{Introduction}
There is variation in the flavor and aroma of different plantation commodities. For example, in Indonesia, the clove buds from Java has a prominent wooden aroma and sour flavor while those in Bali have a sweet-spicy flavor \cite{Broto}. Arabica coffee from Gayo has a lower acidity and a strong bitterness. In contrast, coffee from Toraja has a medium browning, tobacco, or caramel flavor, not too acidic and bitter. Furthermore, Kintamani coffee from Bali has a fruit flavor and acidity, mixed with a fresh flavor. Contrastingly, Coffee from Flores has a variety of flavors ranging from chocolate, spicy, tobacco, strong, citrus, flowers and wood. Coffee from Java has a spicy aroma while that from Wamena has a fragrant aroma and without pulp \cite{coffeland}.

The specific flavors and aromas are attributed to the composition of  commodities' metabolites. Generally, specific metabolite contributes is responsible for particular flavors and aroma. For this reason, it is vital to recognize the characteristics of each plantation commodity based on  metabolite composition. This study investigates the origin of clove buds. This helps to maintain the flavor of a product using clove buds as a mixture. Also, the characteristics of food products can be predicted based on the origin of clove buds used due to differences flavour and taste between regions \cite{kresnowati}.

Metabolic profiling is a widely used approach in obtaining information related to metabolites contained in a biological sample. This is a quantitative measurement of metabolites from biological samples \cite{kopka,putri}. To give meaning to the metabolites data sets, chemometrics technique was developed. This is a chemical sub-discipline that uses mathematics, statistics and formal logic to gain knowledge about chemical systems. It provides maximum relevant information by analysing metabolites data sets from biological samples \cite{massart}. Additionally, it is used in pattern recognition of metabolites data sets in complex chemical systems \cite{kresnowati}. Pattern recognition in biological samples identifies specific metabolites or biomarkers that form particular flavor and aroma.

Artificial neural networks have been widely used in pattern recognition \cite{cornelius} and other applications in various fields as shown by some researchs \cite{samir,25,26,27,28,29}. However, it has not been fully implemented, especially in clove buds. The small data sets available limits the implementation of artificial neural networks for clove buds. This is attributed to the lack of metabolite composition in the clove buds and high cost for extracting them. Furthermore, some clove buds have zero metabolite concentration. However, this is because of inefficient tools in the laboratory to detect metabolites whose values are very small. Therefore this study implements artificial neural networks as pattern recognition in clove buds data sets. Each origin of clove buds has specific metabolites as a biomarker.

\section{Materials and Methods}

\subsection{Materials}
This study uses clove buds data sets obtained from Kresnowati et al. \cite{kresnowati}. It examined clove buds from four origins in Indonesia, including Java, Bali, Manado and Toli-Toli. Each origin has three regions, and therefore, there are twelve regions in total. In the laboratory, eight experiments are carried out in each region, except for Java with only six experiments. Each experiment, 47 types of metabolites were recorded. In the matrix, data sets are 94 $\times$ 47. The row and column represent the number of experiments and metabolites, respectively.

\subsection{Data Preprocessing}
In total, the clove buds data sets have a wide range, specifically between $10^{-4}$ and 10. Therefore, logarithmic transformations are used to obtain reliable numerical data. Since some metabolites data have zero concentration, logarithmic transformation cannot be applied directly. This is because their concentrations range below the specified threshold. The metabolite data with zero concentration are not removed because of acting as biological markers. Therefore, they are replaced with value of one order smaller than the smallest concentration available. In this case, the zeros are replaced with $10^{-5}$. Before implementing artificial neural networks, one stage preprocessing clove buds data sets from \cite{rustam} are added to normalize the values of metabolites data. Normalization  ensure that each  data has the same influence or contribution to determine its origin. The following normalization formula is used \cite{beltramo}

\begin{equation}
z_{kl}=\frac{{x}_{kl}-\overline{x}}{s}.
\end{equation}
Here $z_{kl}$ is the result of normalization of $x_{kl}$, $\overline{x}$ is the mean of the $k$-th experiment and $s$ is

\begin{equation}
s=\sqrt{\sum_{k=1}^{n}\frac{{x}_{kl}-\overline{x}}{n-1}}.
\end{equation}

\subsection{Artificial Neural Network}

Artificial neural networks are a false representation of the human brain that simulates the learning process \cite{fausett}. Backpropagation and resilient propagation are learning algorithms widely used in artificial neural networks \cite{aizenberg,johansson,pleune,kaiser,el2015,chayjan,anastasiadis,santra,patnaik,fisch,shiblee}. In this study, two different network architectures, including resilient and backpropagation are used. The first and second architectures consist of two and one hidden layers, respectively.

\subsubsection{Backpropagation Learning Algorithm}
Backpropagation learning algorithm is based on the repeated use of chain rules to calculate the effect of each weight in  network concerning the error function $E$ \cite{riedmiller}.

\begin{equation}
\frac{\partial E}{\partial w_{ij}}=\frac{\partial E}{\partial o_{i}}\frac{\partial o_{i}}{\partial net_{i}}\frac{\partial net_{i}}{\partial w_{ij}}
\end{equation}
where $w_{ij}$ is the weight from $j-th$ neuron to $i-th$ neuron, $o_{i}$ is the output, and $net_{i}$ is the weighted number of neurons input $i$. Once the partial derivatives for each weight are known, the goal of minimizing the error function is achieved with gradient descent \cite{riedmiller}:

\begin{equation}
w_{ij}^{(t+1)}=w_{ij}^{(t)}-\epsilon \frac{\partial E}{\partial w_{ij}}^{(t)}
\label{eq:4}
\end{equation}
where $t$ is iteration and $0<\epsilon<1$ the learning rate. From the Equation (\ref{eq:4}), choosing a large learning rate (close to 1), allows for oscillations. This makes the error fall above the specified  tolerance value and lessens the identification accuracy. Conversely, in case, the learning rate ($\epsilon$) is too small (close to 0), many steps are needed for  convergence of the error function $E$. To avoid these, the backpropagation learning algorithm is expanded by adding the momentum parameter $(0 <\alpha<1)$ as shown in Equation (\ref{eq:5}). The addition of momentum parameter also accelerates the convergence of error function \cite{riedmiller}.

\begin{equation}
\Delta w_{ij}^{(t+1)}=-\epsilon \frac{\partial E}{\partial w_{ij}}^{(t)} + \alpha \Delta w_{ij}^{(t-1)}
\label{eq:5}
\end{equation}
where it measures the effect of  previous step on the currently.

To activate neurons in the hidden and output layers, the sigmoid activation function is used. Three essential properties used in backpropagation and resilient propagation include bounded, monotonic and continuously differentiable. This helps to convert a weighted amount of input into an output signal for each neuron $i$ as shown by Equation (\ref{eq:6}) \cite{bhagat}.

\begin{equation}
O_{i}=f(I_{i})=\frac{1}{1+e^{-\sigma I_{i}}}.
\label{eq:6}
\end{equation}
where $I_{i}$ is the input of $i$-th weighted number of neuron, $\sigma$ the slope parameter of the sigmoid activation function and $O_{i}$ the output of $i$-th neuron. The threshold used on the output layer for the sigmoid activation function is

\begin{equation}
O_{i} =\left\{\begin{matrix}
	1 ~~ if ~~ O_{i} \geq 0.5\\ 
	0 ~~ if ~~ O_{i} < 0.5
\end{matrix}\right.
\end{equation}
The weighted amount input is given in the following equation \cite{bhagat}.

\begin{equation}
\sum_{i=1}^{n} w_{ij} O_{i}+w_{Bj} O_{B}.
\label{eq:8}
\end{equation}
The sum of $i$ represents the input received from all neurons in the input layer, while $B$ is the bias neuron. Weight $w_{ij}$ is associated with connections from $i$-th neuron to $j$-th neuron, while $w_{Bj}$ weight  relates to the connections from biased to $j$-th neuron. The weighted amount obtained in the hidden  and the output layers are activated by substituting the weighted amount from Equation (\ref{eq:8}) to be an exponent in Equation (\ref{eq:6}).

\subsubsection{Resilient propagation Learning Algorithm}

Riedmiller et al. in \cite{riedmiller} proposed a resilient propagation learning algorithm developed by the backpropagation algorithm. The algorithm directly adapts to the weight value based on local gradient information. Riedmiller et al. \cite{riedmiller} introduced an update value $\Delta_{ij}$ for each weight determining the size of weight update. The adaptive update value evolves during the learning process based on its local sight on the error function $E$, according to the following learning rule \cite{riedmiller}:

\begin{equation}
\Delta_{ij}^{t}=\left\{\begin{matrix}
 \eta^{+} * \Delta_{ij}^{(t-1)},  ~~ if ~~ \frac{\partial E}{\partial w_{ij}}^{(t-1)}*\frac{\partial E}{\partial w_{ij}}^{(t)}>0\\ 
  \eta^{-} * \Delta_{ij}^{(t-1)}, ~~ if ~~ \frac{\partial E}{\partial w_{ij}}^{(t-1)}*\frac{\partial E}{\partial w_{ij}}^{(t)}<0\\
 \Delta_{ij}^{(t-1)}~~~~ ,~~~~~~~~~~~~~~~~~~  else \\ \end{matrix}
\right.
\end{equation}
where ($0<\eta^{-}<1<\eta^{+}$) $\eta^{-}$ and $\eta^{+}$ represents the decrease and increase factors, respectively. According to this adaptation rule, every time the partial derivative of the corresponding weight $w_{ij}$ changes its sign, which indicates that the last update is too big and the algorithm is above the local minimum, the update value $\Delta_{ij}$ is decreased by the factor $\eta^{-}$. In case the derivative retains its sign, the update value  slightly increases to accelerate convergence in the shallow regions \cite{riedmiller}.

Once the update value for each weight is adjusted, the update weight itself follows rule stating that in case the derivative is positive, the weight is decreased by its update value. If the derivative is negative, the update value is added

\begin{equation}
\Delta w_{ij}^{t}=\left\{\begin{matrix}
  -\Delta_{ij}^{t-1}, ~~ if~~  \frac{\partial E}{\partial w_{ij}}^{(t)}>0\\ 
  +\Delta_{ij}^{t-1},  ~~if~~  \frac{\partial E}{\partial w_{ij}}^{(t)}<0\\
   0 ~~~~~, ~~~~~~~  else \\
\end{matrix}\right.
\end{equation}

\begin{equation}
w_{ij}^{t+1} = w_{ij}^{t}+\Delta w_{ij}^{t}
\end{equation}
However, in case the partial derivative sign changes, which means the previous step was too large and the minimum missed, the previous weight update is reverted:

\begin{equation}
\Delta w_{ij}^{(t)}=-\Delta w_{ij}^{(t-1)},~~ if ~~ \frac{\partial E}{\partial w_{ij}}^{(t-1)}*\frac{\partial E}{\partial w_{ij}}^{(t)}<0
\end{equation} 
Due to the 'backtracking' weight step, the derivative should change its sign once again in the next step. To avoid another problem, there should be no adaptation of update value in the succeeding step. In practice, this can be carried out by setting $\frac{\partial E}{\partial w_{ij}}^{(t-1)}$ = 0 in the $\Delta_{ij}$ adaptation rule. The update values and weights are changed every time the whole set of patterns is presented once to the network (learning by epoch).

The following shows the process of adaptation and resilient propagation learning process. The $\mathbf{minimum (maximum)}$ operator is expected to provide a minimum or maximum of two numbers. The sign operator returns +1 in the argument is positive, -1 in case the it is negative, and 0 for otherwise.
\begin{equation}
\begin{split}
For~each ~weight~ and ~bias\{\\
\mathbf{if}(\frac{\partial E}{\partial w_{ij}}^{(t-1)}*\frac{\partial E}{\partial w_{ij}}^{(t)}>0)\mathbf{\, then\{}\\ 
\Delta_{ij}^{(t)}=\mathbf{minimum}(\Delta_{ij}^{(t-1)}*\eta^{+},\Delta_{\mathit{\mathrm{max}}})\\
\Delta w_{ij}^{(t)}=\mathbf{sign}(\frac{\partial E}{\partial w_{ij}}^{(t)}*\Delta_{ij}^{(t)})\\
w_{ij}^{(t+1)}=w_{ij}^{(t)}+\Delta w_{ij}^{(t)}\}\\ 
\mathbf{else \: if}(\frac{\partial E}{\partial w_{ij}}^{(t-1)}*\frac{\partial E}{\partial w_{ij}}^{(t)}<0)\mathbf{\, then\{} \\ 
\Delta_{ij}^{(t)}=\mathbf{maximum}(\Delta_{ij}^{(t-1)}*\eta^{-},\Delta_{\mathit{\mathrm{min}}}) \\ 
w_{ij}^{(t+1)}=w_{ij}^{(t)}-\Delta w_{ij}^{(t-1)}\\ 
\frac{\partial E}{\partial w_{ij}}^{(t)}=0\}\\
\mathbf{if}(\frac{\partial E}{\partial w_{ij}}^{(t-1)}*\frac{\partial E}{\partial w_{ij}}^{(t)}=0)\mathbf{\, then\{} \\  
\Delta w_{ij}^{(t)}=-\mathbf{sign}(\frac{\partial E}{\partial w_{ij}}^{(t)}*\Delta_{ij}^{(t)})\\ 
w_{ij}^{(t+1)}=w_{ij}^{(t)}+\Delta w_{ij}^{(t)}\}\\ 
\}
\end{split}
\end{equation}

\section{Results and Discussions}

In this study, the percentage of training and testing data sets are 80\% and 20\%, respectively. The metabolites data sets in matrix are 94$\times$47. Out of 94 rows, 75 were chosen randomly as training data sets, while the remaining were used as testing. The selection of training data sets is carried out randomly 30 times. Therefore, in each network architecture, there are 30 values for the percentage of identification accuracy, coefficient of determination and the mean squared error ($MSE$). The average is chosen as a representative of the 30 values. In each network architecture, learning rate ($\epsilon$) 0.9, momentum parameter ($\alpha$) 0.1 and maximum epoch 5000 are used with error target $10^{-3}$. In this study, each origin is represented by a binary code. Specifically, the binary code for the  Java origin is 1000, Bali 0100, Manado 0010 and Toli-Toli 0001. The calculation of identification accuracy and $MSE$ is shown in Equation (\ref{eq:13}) and (\ref{eq:14}).

\begin{equation}
\% \: accuracy = \frac{a}{k}100\%
\label{eq:13}
\end{equation}
Where $a$ is the number of origins identified correctly, while $k$ is the total number. $MSE$ calculated by the following equation \cite{bhagat}

\begin{equation}
MSE = \frac{1}{m\cdot n} \sum_{p=1}^{m} \sum_{k=1}^{n}(T_{kp}-O_{kp})^{2}.
\label{eq:14}
\end{equation}
where $T_{kp}$ is the desired target, $O_{kp}$ the network output and $p$ the variable corresponding to the number of origins.

The suitability between the expected target and network output was evaluated based on the coefficient of determination $R^{2}$. It was calculated using the following equation \cite{el2015} 
\begin{equation}
R^{2}=1-\frac{\frac{1}{n} \sum_{k=1}^{n} (T_{kp}-O_{kp})^{2}}{\frac{1}{n-1} \sum_{k=1}^{n} (T_{kp}-\overline{T_{kp}})^{2}}.
\end{equation}
Where $\overline{T_{kp}}$ is the average desired target.

In this study, backpropagation and resilient propagation were used, each consisting of two and one hidden layers. For one hidden layer, the number of neurons was determined using the formula proposed by Shibata and Ikeda in 2009 \cite{shibata}, specifically $N_{h} = \sqrt{N_{i} \cdot N_{o}}$, where $N_{h}$, $N_{i}$, and $N_{o}$ represents hidden, input and output neurons, respectively. In both backpropagation and resilient propagation, the number of neurons used do not exceed one hidden layer. Based on Shibata and Ikeda \cite{shibata} formula, the number of neurons in one hidden layer was obtained, specifically $N_{h} = \sqrt{N_{i} \cdot N_{o}} = \sqrt{47 \cdot 4 } = 13.71$. However, in this study, it was rounded up to 15 neurons. Some experiments were conducted to evaluate the identification accuracy, and whether using one hidden layer 15 neurons might lead to a better accuracy of identification than two hidden layers. However, the number of neurons varied, setting less than 15 neurons. For two hidden layers, experiments were conducted with the number of consecutive neurons as follows; 3-5 (8), 4-6 (10), 5-7 (12) and 6-8 (14). The number of neurons in the hidden layer never exceed 15 neurons.

\subsection{Backpropagation (B-Prop) with Two Hidden Layers}
\label{section:section31}

In this section, backpropagation learning algorithm with two hidden layers was used. The number of neurons in the hidden layer varied with not more than 15 neurons. There were four variations of network architecture, including 47-3-5-4, 47-4-6-4, 47-5-7-4 and 47-6-8-4. The input layer consists of 47 neurons based on the number of metabolites. The output layer consists of 4 neurons according to the number of clove buds origins.

\begin{table*}
	\caption{Backpropagation with two hidden layers}
	\centering
	\begin{tabular}{ccccccc}
		\hline
		\textbf{Network} &  \multicolumn{2}{c}{\textit{\textbf{MSE}}} & \multicolumn{2}{c}{\textbf{Accuracy} (\%)} & \multicolumn{2}{c}{$R^2$} \\
		\cline{2-7}
		\textbf{Architecture} & \textbf{Training} & \textbf{Testing} &  \textbf{Training} & \textbf{Testing} & \textbf{Training} & \textbf{Testing} \\
		\hline
		\textbf{47-3-5-4} & \textbf{0.10346} & \textbf{0.11357} & \textbf{76.98} & \textbf{73.68} & \textbf{0.81}
		& \textbf{0.76} \\
		\hline
		47-4-6-4 & 0.13084	& 0.13547 & 62 & 57.54 & 0.64
		& 0.61 \\
		\hline
		47-5-7-4 & 0.14889	& 0.15884 & 49.73 & 41.4 & 0.51 & 0.42 \\
		\hline
		47-6-8-4 & 0.15388 & 0.15874 & 50.04 & 42.46 & 0.48 & 0.43 \\
		\hline
	\end{tabular}
	\label{tab:tabel1}
\end{table*}

Table \ref{tab:tabel1} shows the network architecture \textbf{47-3-5-4} gives the highest value for identification accuracy and coefficient of determination in training and testing data sets. Similar to the $MSE$, this network architecture provides the smallest amount of both training and the testing data sets. From Table \ref{tab:tabel1}, increasing the number of neurons in the backpropagation with two hidden layers decreases network performance. This is in line with Shafi et al. in 2006 \cite{shafi}, which stated that increasing the number of neurons in the hidden layer only heightened the complexity of the network. Still, it does not increase the accuracy of pattern recognition.

\subsection{Backpropagation (B-Prop) with One Hidden Layer}
\label{section:section32}

The backpropagation learning algorithm with one hidden layer was implemented to evaluate its result in case of a comparison using two hidden layers. The results obtained are shown in Table \ref{tab:tabel2}.

\begin{table*}
	\caption{Backpropagation with one hidden layer}
	\centering
	\begin{tabular}{ccccccc}
		\hline
		\textbf{Network} &  \multicolumn{2}{c}{\textit{\textbf{MSE}}} & \multicolumn{2}{c}{\textbf{Accuracy} (\%)} & \multicolumn{2}{c}{$R^2$} \\
		\cline{2-7}
		\textbf{Architecture} & \textbf{Training} & \textbf{Testing} &  \textbf{Training} & \textbf{Testing} & \textbf{Training} & \textbf{Testing} \\
		\hline
		47-15-4 & 0.0721 & 0.0773 & 99.91 & 99.47 & 0.99 & 0.98 \\
		\hline
	\end{tabular}
	\label{tab:tabel2}
\end{table*}

Table \ref{tab:tabel2} shows that network architecture 47-15-4 identifies the clove buds origin effectively. The identification accuracy percentage is 99.91\% and 99,47\% for training and testing data sets, respectively. Besides, the $MSE$ value is also smaller compared to the two hidden layers. 

For the backpropagation algorithm, the results show one hidden layer is better than two. This is in line with Villiers and Barnard \cite{de1993}, which stated that network architecture with one hidden layer is on average better than two hidden layers. They concluded that two hidden layers are more difficult to train. Additionally, they also established that this behaviour is caused by a local minimum problem. The networks with two hidden layers are more prone to the local minimum problem during training.  

\subsection{Resilient Propagation (R-Prop) with Two Hidden Layers}

Resilient propagation learning algorithm contains some parameters, including the upper and lower limits, as well as the decrease and increase factors. In this study, the range of update values is limited to the upper limit ($\Delta_{max}$) = 50, the lower limit ($\Delta_{min}$) = $10^{-6}$, and the decrease and increase factors ($\eta^{-}$) = 0.5 and ($\eta^{+}$) = 1.2, respectively. The reason for choosing these values is shown in \cite{riedmiller}.
Similar to Section \ref{section:section31}, the resilient propagation learning algorithm is applied to the network architecture with two hidden layers. The number of neurons vary but do not exceed 15 neurons. In this section, there are four variations of network architecture, including 47-3-5-4, 47-4-6-4, 47-5-7-4 and 47-6-8-4. 

\begin{table*}
	\caption{Resilient propagation with two hidden layers}
	\centering
	\begin{tabular}{ccccccc}
		\hline
		\textbf{Network} & \multicolumn{2}{c}{\textit{\textbf{MSE}}} & \multicolumn{2}{c}{\textbf{Accuracy} (\%)} & \multicolumn{2}{c}{$R^2$} \\
		\cline{2-7}
		\textbf{Architecture} & \textbf{Training} & \textbf{Testing} &  \textbf{Training} & \textbf{Testing} & \textbf{Training} & \textbf{Testing} \\
		\hline
		47-3-5-4 & 0.07209 & 0.08111 & 99.73 & 97.37 & 0.98 & 0.95 \\
		\hline
		47-4-6-4 & 0.07162 & 0.08316 & 99.69 & 96.49 & 0.99 & 0.94 \\
		\hline
		\textbf{47-5-7-4} & \textbf{0.07160} & \textbf{0.07961} & \textbf{99.96} & \textbf{97.89} & \textbf{0.99} & \textbf{0.96} \\
		\hline
		47-6-8-4 & 0.07160 & 0.07978 & 99.78 & 97.72 & 0.99 & 0.96 \\
		\hline
	\end{tabular}
	\label{tab:tabel3}
\end{table*}

The results in Table \ref{tab:tabel3} show the network architecture \textbf{47-5-7-4} gives the highest identification accuracy of clove buds origin. The percentage of identification accuracy is 99.96\% and 97.89\% for training and testing data sets, respectively.

\subsection{Resilient Propagation (R-Prop) with One Hidden Layer}

In this section, the resilient propagation learning algorithm is implemented with one hidden layer. Similar to section \ref{section:section32}, the number of neurons in the hidden layer is 15 neurons, and have the network architecture 47-15-4. 

\begin{table*}
	\caption{Resilient propagation with one hidden layer}
	\centering
	\begin{tabular}{ccccccc}
		\hline
		\textbf{Network} & \multicolumn{2}{c}{\textit{\textbf{MSE}}} & \multicolumn{2}{c}{\textbf{Accuracy} (\%)} & \multicolumn{2}{c}{$R^2$} \\
		\cline{2-7}
		\textbf{Architecture} & \textbf{Training} & \textbf{Testing} &  \textbf{Training} & \textbf{Testing} & \textbf{Training} & \textbf{Testing} \\
		\hline
		47-15-4 & 0.07158 & 0.07932 & 99.86 & 94.74 & 0.99 & 0.92 \\
		\hline
	\end{tabular}
	\label{tab:tabel4}
\end{table*}

Table \ref{tab:tabel4} shows the network architecture 47-15-4 identifies the origin of clove buds with identification accuracy of 99.86\% and 94.74\% on training and testing data sets, respectively. 

The network architecture of the resilient propagation algorithm, both two hidden layers and one hidden layer, provides identification results with very high accuracy. However, the network architecture with two hidden layers is slightly lower accuracy.

Tables \ref{tab:tabel3} and \ref{tab:tabel4} show the two-layered resilient propagation with deficient neurons performs better than the single-layer having more neurons. This is in line with Santra et al. \cite{santra}, established that the performance of two hidden layers with 8-10 (18) neurons is better than one hidden layer with 62 neurons.

\begin{figure}[h!]
	\centering
	\includegraphics[width=7cm,height=6cm]{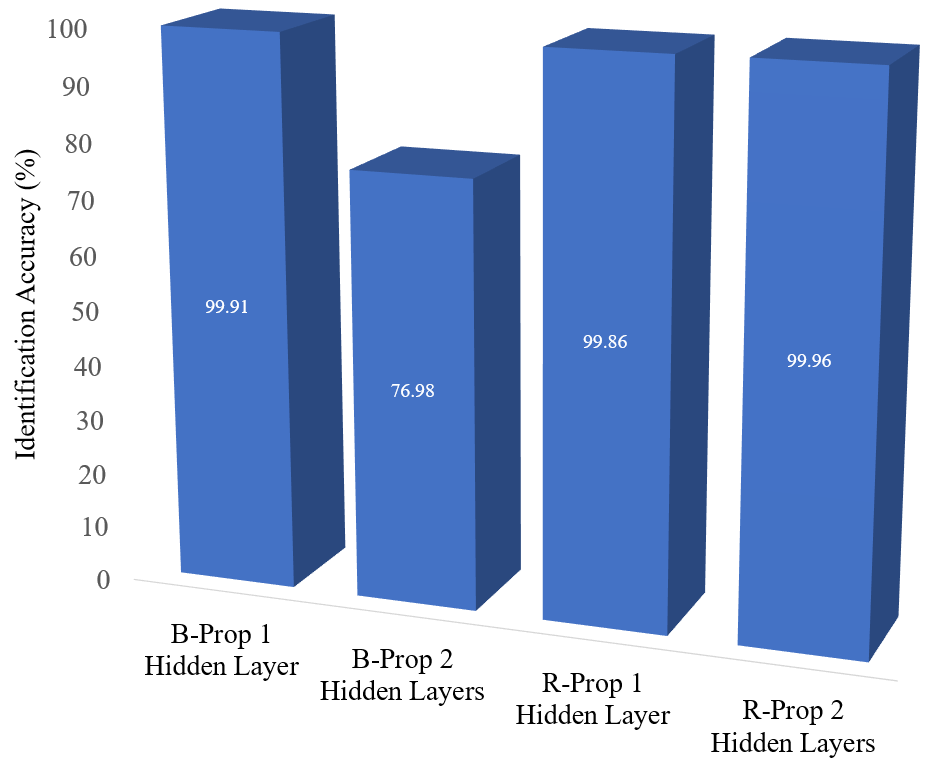} % <-- use this for your graphics
	\caption{Identification accuracy percentage of training data sets.}
	\label{fig:training}
\end{figure}

\begin{figure}[h!]
	\centering
	\includegraphics[width=7cm,height=6cm]{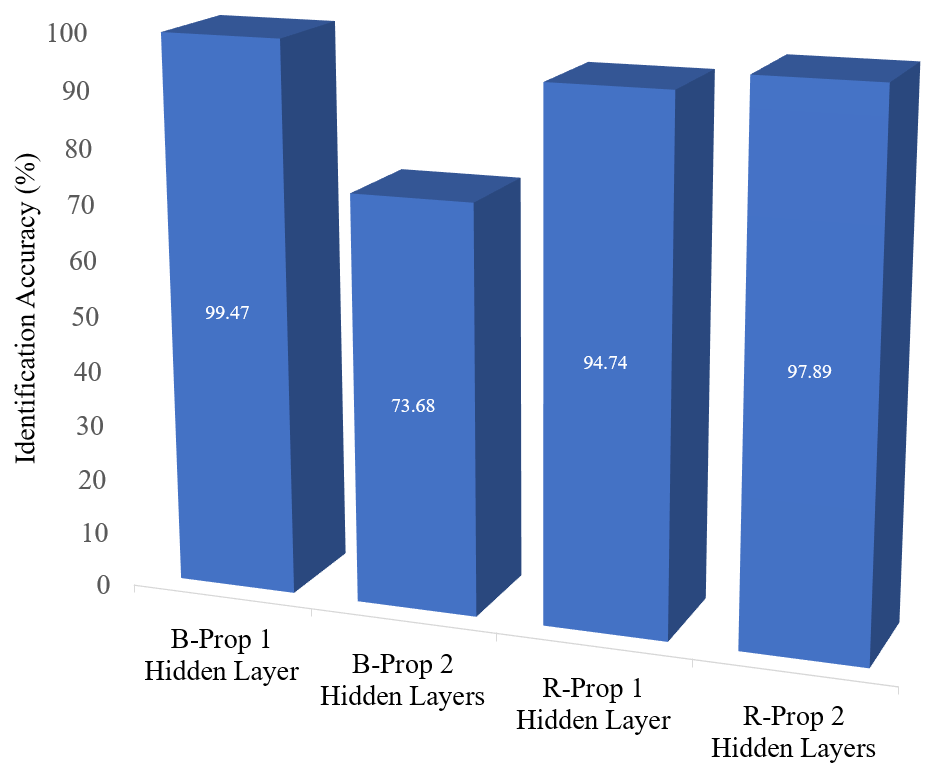} % <-- use this for your graphics
	\caption{Identification accuracy percentage of testing data sets.}
	\label{fig:testing}
\end{figure}

\begin{figure}[h!]
	\centering
	\includegraphics[width=7cm,height=6cm]{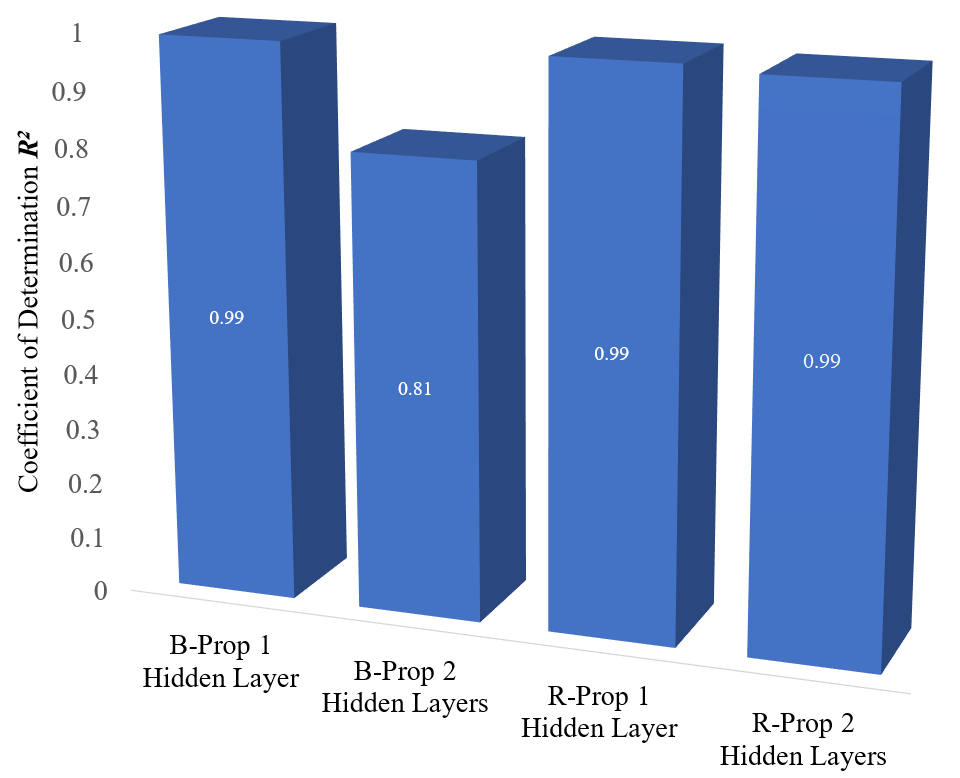} % <-- use this for your graphics
	\caption{Determination coefficient of training data sets.}
	\label{fig:r_training}
\end{figure}

\begin{figure}[h!]
	\centering
	\includegraphics[width=7cm,height=6cm]{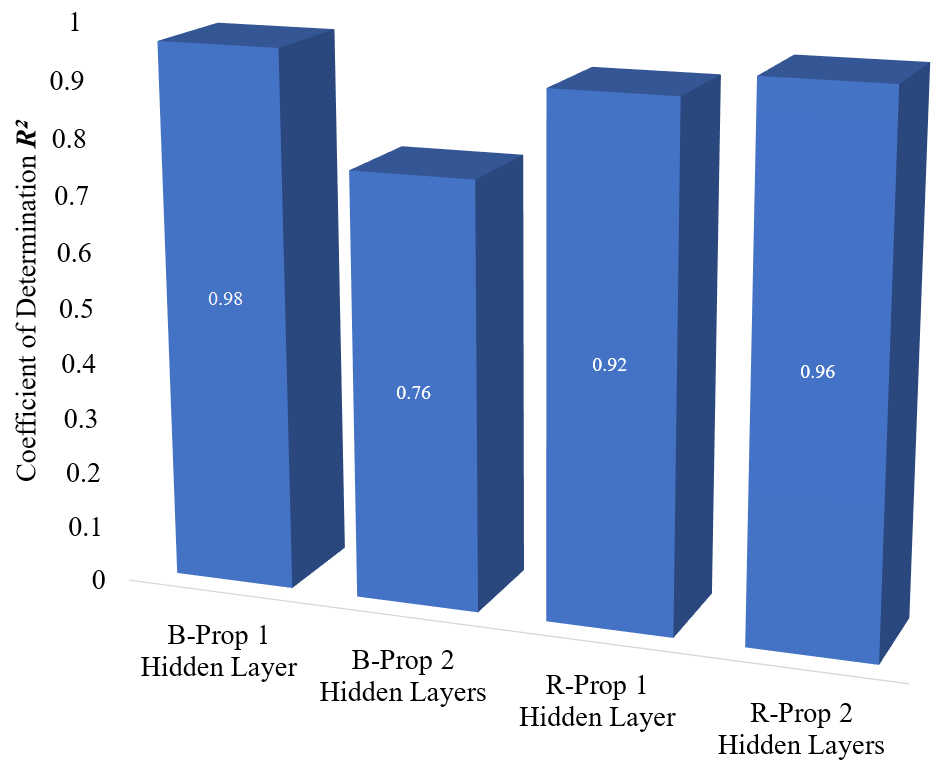} % <-- use this for your graphics
	\caption{Determination coefficient of testing data sets.}
	\label{fig:r_testing}
\end{figure}

\begin{figure}[h!]
	\centering
	\includegraphics[width=7cm,height=6cm]{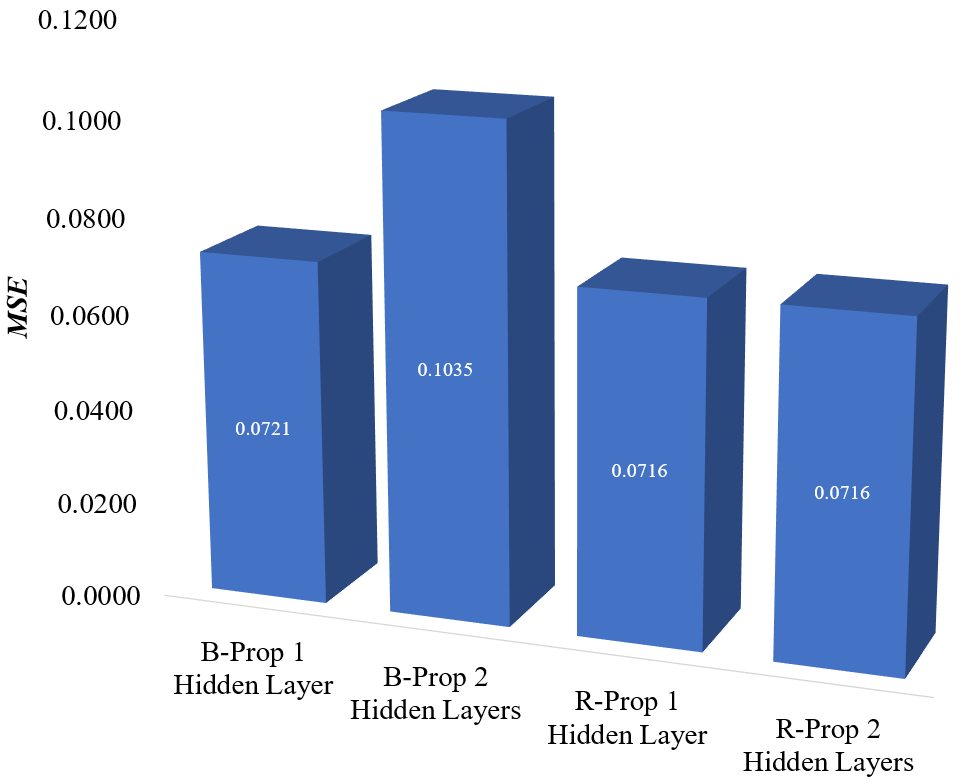} % <-- use this for your graphics
	\caption{$MSE$ of training data sets.}
	\label{fig:mse_training}
\end{figure}

\begin{figure}[h!]
	\centering
	\includegraphics[width=7cm,height=6cm]{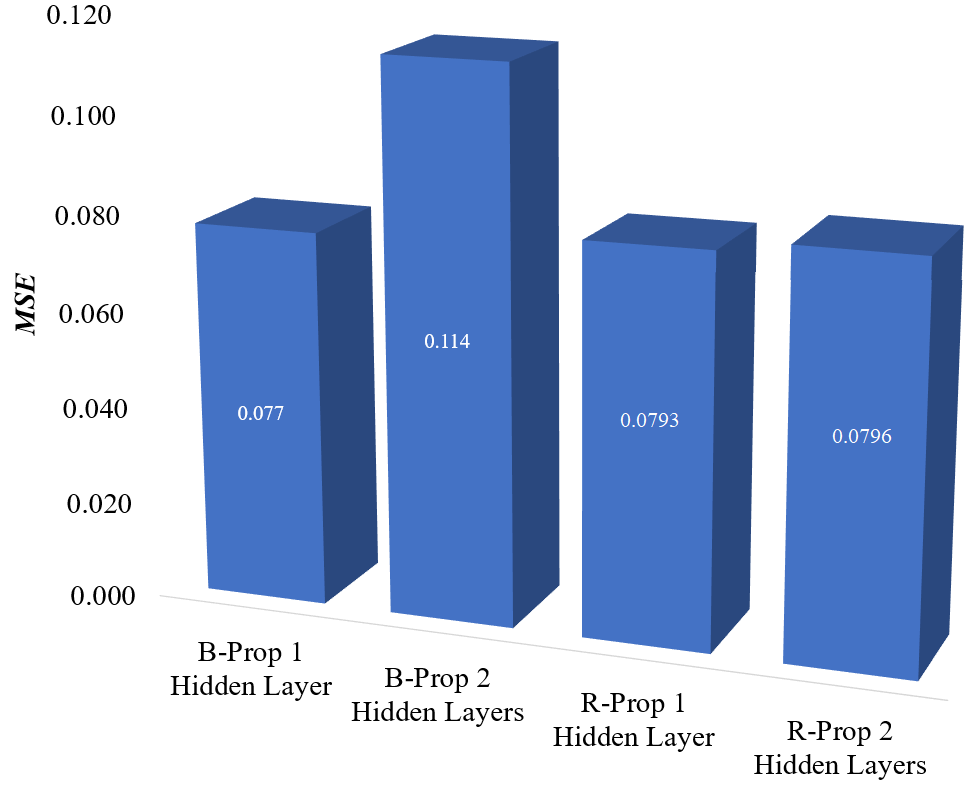} % <-- use this for your graphics
	\caption{$MSE$ of testing data sets.}
	\label{fig:mse_testing}
\end{figure}

The summary of the best identification accuracy and determination coefficient are shown in Figures \ref{fig:training}, \ref{fig:testing}, \ref{fig:r_training} and \ref{fig:r_testing}, respectively. For each network architecture, the smallest \textit{MSE} in training and testing data sets are shown in Figures \ref{fig:mse_training} and \ref{fig:mse_testing}, respectively.

%The smallest \textit{MSE} in the training data sets obtained from the resilient propagation algorithm, both one and two hidden layers (see Figure \ref{fig:mse_training}). While in the testing data sets, the backpropagation algorithm with one hidden layer gives the smallest \textit{MSE} (see Figure \ref{fig:mse_testing}).

The results of identification from the origins of clove buds have been obtained. In small data set categories, backpropagation with one hidden layer provides accurate identification in the training and testing data sets. It accurately identifies the origins of clove buds obtained using the resilient propagation algorithm with two hidden layers.

The neural networks model obtained in this paper can be a reference from a scientific perspective. For instance, it can be used in future studies to identify the origin of various plantation commodities with small metabolites data sets. At the moment, the most appropriate way of determining the origin of a plantation commodity is qualitative, relying on the services of flavorist to evaluate flavor and taste. This is because each commodity has a specific flavor and taste based on the origin of its region. Furthermore, the different origins of clove buds data sets have not been reported in the literature and thus no direct comparison can be presented in this paper. 

\section{Conclusions}

This paper demonstrated the potential and ability of a neural network approach with backpropagation and resilient propagation learning algorithms. It was meant to identify the clove buds origin based on metabolites composition. The work was divided into two parts, the first one being identification of the clove buds origin using the backpropagation learning algorithm. Two network architectures were constructed, each containing one hidden and two hidden layers. The results showed the use of one hidden layer gives clove buds origin identification accurately, specifically 99.91\% and 99.47\% in training and testing data sets. The second step involved the identification of the clove buds origin using  resilient propagation learning algorithm. Two network architectures were constructed, each containing one hidden and two hidden layers. The results showed the use of two hidden layers gives accurate clove buds origin identification, including 99.96\% and 97.89\% in training and testing data sets. From these results, it was concluded that for identification of a small of metabolites data sets from a plantation commodity, the backpropagation algorithm with one hidden layer and the resilient propagation algorithm with two hidden layers should be used. This paper also confirmed the contribution of artificial neural networks to pattern recognition of metabolites data sets obtained by metabolic profiling technique.

%\begin{acknowledgements}
%The authors express gratitude to the government of Indonesia, especially Endowment Fund for Education (LPDP - Lembaga Pengelola Dana Pendidikan), for its funding.
%\end{acknowledgements}

\bibliographystyle{unsrt}
%\bibliography{references}  %%% Remove comment to use the external .bib file (using bibtex).
%%% and comment out the ``thebibliography'' section.

\bibliography{references}

\end{document}